\documentclass[a4paper,fleqn]{cas-dc}

\usepackage[authoryear,longnamesfirst]{natbib}

\def\tsc#1{\csdef{#1}{\textsc{\lowercase{#1}}\xspace}}
\tsc{WGM}
\tsc{QE}

\begin{document}
\let\WriteBookmarks\relax
\def\floatpagepagefraction{1}
\def\textpagefraction{.001}

\shorttitle{BCGGAN: Ballistocardiogram artifact removal in simultaneous EEG-fMRI using generative adversarial network}    

\shortauthors{Lin, Zhang, Liu et al.}  

\title [mode = title]{BCGGAN: Ballistocardiogram artifact removal in simultaneous EEG-fMRI using generative adversarial network}



%

\author[1,2]{Guang Lin}[orcid=0000-0002-4948-1496]


\author[1,2]{Jianhai Zhang}[]
\cormark[1]
\author[1,2]{Yuxi Liu}[]
\author[1,2]{Tianyang Gao}[]
\author[1,2]{Wanzeng Kong}[]
\author[3,4]{Xu Lei}[]
\author[5]{Tao Qiu}[]


\affiliation[1]{organization={HangZhou Dianzi University},
            city={HangZhou},
            postcode={310018}, 
            country={China}}
\affiliation[2]{organization={Key labortory of Brain Machine Collaborative Intelligence of Zhejiang Province},
            city={HangZhou},
            postcode={310018}, 
            country={China}}
\affiliation[3]{organization={Southwest University},
            city={Chongqing},
            postcode={400715}, 
            country={China}}
\affiliation[4]{organization={Key Laboratory of Cognition and Personality},
            city={Chongqing},
            postcode={400715}, 
            country={China}}
\affiliation[5]{organization={Zhejiang Provincial Hospital of Chinese Medicine},
            city={HangZhou},
            postcode={310006}, 
            country={China}}

\begin{abstract}
Due to its advantages of high temporal and spatial resolution, the technology of simultaneous electroencephalogram-functional magnetic resonance imaging (EEG-fMRI) acquisition and analysis has attracted much attention, and has been widely used in various research fields of brain science. However, during the fMRI of the brain, ballistocardiogram (BCG) artifacts can seriously contaminate the EEG. As an unpaired problem, BCG artifact removal now remains a considerable challenge. Aiming to provide a solution, this paper proposed a novel modular generative adversarial network (GAN) and corresponding training strategy to improve the network performance by optimizing the parameters of each module. In this manner, we hope to improve the local representation ability of the network model, thereby improving its overall performance and obtaining a reliable generator for BCG artifact removal. Moreover, the proposed method does not rely on additional reference signal or complex hardware equipment. Experimental results show that, compared with multiple methods, the technique presented in this paper can remove the BCG artifact more effectively while retaining essential EEG information.
\end{abstract}



\maketitle

\begin{sloppypar}
\section{Introduction}
\label{Introduction}
The simultaneous application of electroencephalography (EEG) and functional magnetic resonance imaging (fMRI) enables both high temporal and spatial resolution when measuring brain activity (\cite{mulert2009eeg}). This non-invasive and safe neuroimaging technique has proven valuable in clinical diagnosis and cognitive neuroscience research (\cite{hsiao2018neurophysiological,yang2018exploration,tong2019real}), such as epileptic event localization (\cite{hosseini2020multimodal,hur2020guideline}) and brain state analysis (\cite{hunyadi2019dynamic,mash2020atypical}).

Despite its many successful applications, the utility of EEG-fMRI is still fundamentally limited by various artifacts introduced by the mutual influence of the two types of equipment during simultaneous acquisition. The artifacts introduced by the EEG system into the fMRI data are minor, and do not significantly affect the fMRI quality. However, the fMRI system can bring severe artifacts into the EEG signals, mainly of two types (\cite{debener2007improved}). One major artifact is called gradient artifact (GA), which repeats in a stereotypical way (\cite{niazy2005removal}) and is caused by the rapid switching of magnetic field during fMRI acquisition. This artifact has a large amplitude, therefore, due to its time-shift invariance, it can be successfully removed from EEG signals using average artifact subtraction (AAS) (\cite{allen1998identification}) or optimal basis set (OBS) (\cite{niazy2005removal,wu2016real}). The other significant artifact type, which is the focus of this paper, is the ballistocardiogram (BCG) artifact, which is caused by various factors such as minute head movements caused by the heartbeat and the expansion and contraction of the head's blood vessels with the heartbeat under high-intensity magnetic field conditions (\cite{nakamura2006removal,debener2008properties}). The BCG artifact is non-stationary, varies from subject to subject and channel to channel, and changes over time (\cite{marino2018adaptive}). Therefore, the removal of this artifact type is a crucially important and challenging task.

Many previous studies have focused on BCG artifact removal, with significant developments achieved in average artifact subtraction (AAS)-based methods (\cite{allen1998identification,mullinger2013identifying,steyrl2018online}), principal component analysis (PCA)-based methods (\cite{niazy2005removal,marino2018adaptive,allen2000method}), and independent component analysis (ICA)-based methods (\cite{srivastava2005ica,ghaderi2010removal,vanderperren2010removal,barros1998removing}). The AAS-based methods rely on the repetitive pattern of the BCG artifact, and generate an artifact template to subtract it from the EEG signals. In order to make the template adapt to the change of BCG artifact, the dynamic average artifact template was developed based on median-filtering or Gaussian weighted averaging (\cite{ellingson2004ballistocardiogram,laufs2012personalized}). Most AAS-based methods need a reference electrocardiogram (ECG) signal to generate the template (\cite{steyrl2017reference}), whereas the BCG is non-stationary and propagates in a non-linearly dependent manner on the ECG (\cite{marino2018adaptive}), making it difficult to precisely reconstruct the BCG artifact using ECG. PCA and ICA, which are representative methods of blind source separation, separate the original EEG signals into different components, identify BCG artifact components, and finally remove them. The primary challenge with both methods is the definition of a consistent standard for artificial component selection (\cite{leclercq2009rejection,liu2012statistical}): a small number of sources may leave some artifacts in the signal, whereas a large number of sources may eliminate vital EEG information. Besides, professional hardware equipment, such as carbon fiber sling, are also used to measure the BCG waveform, subtracting it from the contaminated EEG signals. This method can fundamentally solve the problem of BCG artifact, whereas these equipments are often highly expensive, complicated to operate, and are time consuming to prepare (\cite{xia2013bcg,chowdhury2014reference,luo2014ballistocardiogram,van2016carbon}).

Deep learning has received a lot of attention and has been successfully applied in image and audio denoising (\cite{abouzid2019signal,tian2020deep}), and good application on EEG measurements (\cite{aydin2019deep}). Certain studies in recent years (\cite{yang2018automatic,mcintosh2020ballistocardiogram}) used deep learning methods to solve artifact problems. \cite{yang2018automatic} proposed a novel method based on a deep learning network (DLN) to remove ocular artifacts (OA) from contaminated EEG. Therein, training samples without OA are intercepted in the training phase and used to train a DLN to reconstruct the EEG signals. In the testing phase, the trained DLN is used as a filter to automatically remove OA from the contaminated EEG signals. \cite{mcintosh2020ballistocardiogram} presented a novel method for BCG artifact suppression using recurrent neural networks (RNNs). This method depends on the ECG reference signal and restores the EEG through the nonlinear mappings between ECG and BCG-corrupted EEG. At the same time, the authors acknowledge the limitation that they have not compared their results to data recorded outside the range of the static magnetic field, and did not verify the Eyes-Open/Eyes-Closed (EO/EC) effect due to the lack of dataset.

During the application of deep learning, denoising tasks are more complicated than classification tasks. A classification task is often a paired data-to-data problem, which is to learn the mapping between input data X and output data Y using a training set of aligned data pairs. However, for the denoising task, it is hard to directly obtain the paired data for training. In BCG artifact removal, it is impossible to simultaneously obtain BCG-corrupted EEG data and clean EEG data in the same state. When addressing the image denoising problem, the noisy dataset X can be generated by adding various noises to the original dataset Y, thereby creating paired training data. However, since the noise and artifacts in EEG are more complex, this method is not feasible. Thus, converting BCG-corrupted EEG data into clean EEG data constitutes an unpaired signal-to-signal problem. \cite{zhu2017unpaired} presented a novel generative adversarial network (GAN) model (\cite{goodfellow2014generative}) for learning to translate an image from a source domain X to a target domain Y in the absence of paired examples, which provides us with an essential idea for removing the BCG artifact.

This paper uses the GAN to learn how to transform a signal from the BCG-corrupted EEG data to the clean EEG data without paired examples. The GAN is based on two assumptions: a strong discriminator can be built to distinguish the data features of source domain and target domain; the other is that there is a reliable generator that can transform the original features and reconstruct the target features. Since there are huge differences between the two types of data mentioned above, establishing a strong discriminator to distinguish the two types of data is straightforward. Nevertheless, regarding the second assumption, a reliable generator is not easy to obtain due to the low signal-to-noise ratio (SNR) of EEG signals. In order to solve the above problems to more effectively remove the BCG artifact and retain the original EEG signal information to the greatest extent, this paper has made the following contributions:

\begin{itemize}
\item A novel GAN-based model was designed (BCGGAN) to remove the BCG artifact in simultaneous EEG-fMRI. This model does not require additional hardware or reference signal, such as carbon fiber sling or ECG signals.
\item A modular training strategy was proposed to optimize the generator network. The strategy modularizes the generator network and adds additional constraints to train the modules. By this means, better independent module parameters can be obtained to improve the performance of the entire generator network model.
\end{itemize}

Finally, it was confirmed that BCGGAN is effective across multiple evaluations, and has better performance than various methods include Analyzer-2 (a commercial software used to remove the BCG artifact).

\begin{figure*}[ht]
\includegraphics[width=\textwidth]{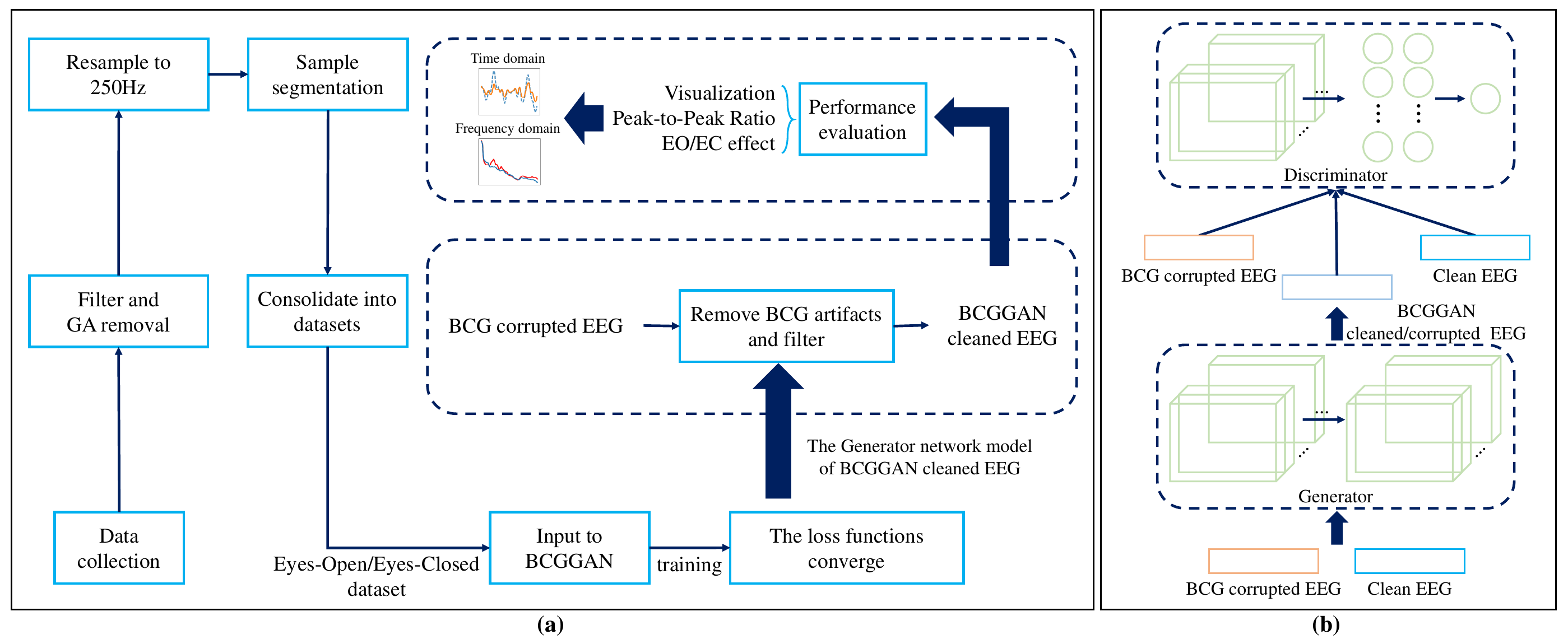}
\caption{Processing pipeline and model architecture. (a) The overall structure of methods and experiments. (b) The GAN framework in CycleGAN.}
\label{FIG:1}
\end{figure*}

\section{Materials and Methods}
\label{Materials and Methods}
Fig.~\ref{FIG:1}a shows the overall framework of our work and how data is transformed in the GAN.

\subsection{Experimental Paradigm}
\label{Experimental paradigm}
A total of five healthy volunteers (2 women) with a mean age of 29 years (range 23–33 years) participated in the experiment. They were all healthy volunteers with no history of neurological disorders, who gave their written informed consent and were paid for their participation. The study was approved by the local ethics committee. 

Once the subjects were put on the EEG caps, they were asked to perform the tasks of Eyes-Open and Eyes-Closed twice, once recording out of the scanner room and once into the scanner. Each task included collecting 5 min of EEG data for Eyes-Open and Eyes-Closed resting states, respectively.

For the acquisition of EEG data, a 32-channel MRI-compatible Brain Products system was utilized for EEG recording (BrainAmp MR plus, Brain Products, Munich, Germany), consisting of 30 scalp electrodes and two additional ECG and EMG electrodes. In the BCGGAN model design, only 30 scalp electrodes are used. The EEG data was collected according to the international 10-20 system with reference electrode at FCz and a sampling frequency of 5 kHz. The SyncBox (SyncBox MainUnit, Brain Products GmbH, Munich, Germany) was used in the synchronous amplifier system to ensure the temporal stability of EEG data acquisition in scanning. The amplified EEG data was transmitted to the computer outside the scanner room for recording.

\subsection{EEG Preprocessing}
\label{EEG preprocessing}
Data preprocessing, as well as actual artifact removal were performed in the MATLAB 9.2.0 (R2017a) (The Mathworks Inc., Natick, Massachusetts, USA) environment with the EEGLAB v14.1.1 toolbox (\cite{delorme2004eeglab}).

All data acquired through the fMRI scanner were subjected to the following preprocessing steps. First, a band-pass filter from 0.1 Hz to 70 Hz was applied to remove low- and high-frequency noise. Next, gradient artifact was removed by using the fMRI artifact slice template removal (FASTR) method (\cite{niazy2005removal,iannetti2005simultaneous}) implemented in EEGLAB, provided by the University of Oxford Centre for Functional MRI of the Brain (FMRIB).

The signals recorded outside the scanner room were also used in this paper to provide a reference, as such acquisitions do not have a BCG artifact. Some simple operations were performed to remove eye artifacts, muscle artifacts, and baseline drift. To save memory, all data were downsampled to 250 Hz. Subsequently, the processed data could be used directly for deep learning methods.

Next, the preprocessed data were divided into samples and integrated into datasets. Although the BCG artifact does not appear periodically, their frequency of occurrence is high. The appearance period of BCG is less than 1 second, thus 1 s was chosen as the length of time window during data segmentation. A series of slices of size $\boldsymbol{C*L}$ ($\boldsymbol{C}$=30, $\boldsymbol{L}$=250) were obtained, where $\boldsymbol{C}$ represents the number of channels, and $\boldsymbol{L}$ represents the data length. Finally, Eyes-Closed EEG data were integrated into one dataset and Eyes-Open EEG data were integrated into another dataset.

\subsection{Model}
\label{Model}
\begin{figure*}[ht]
\includegraphics[width=\textwidth]{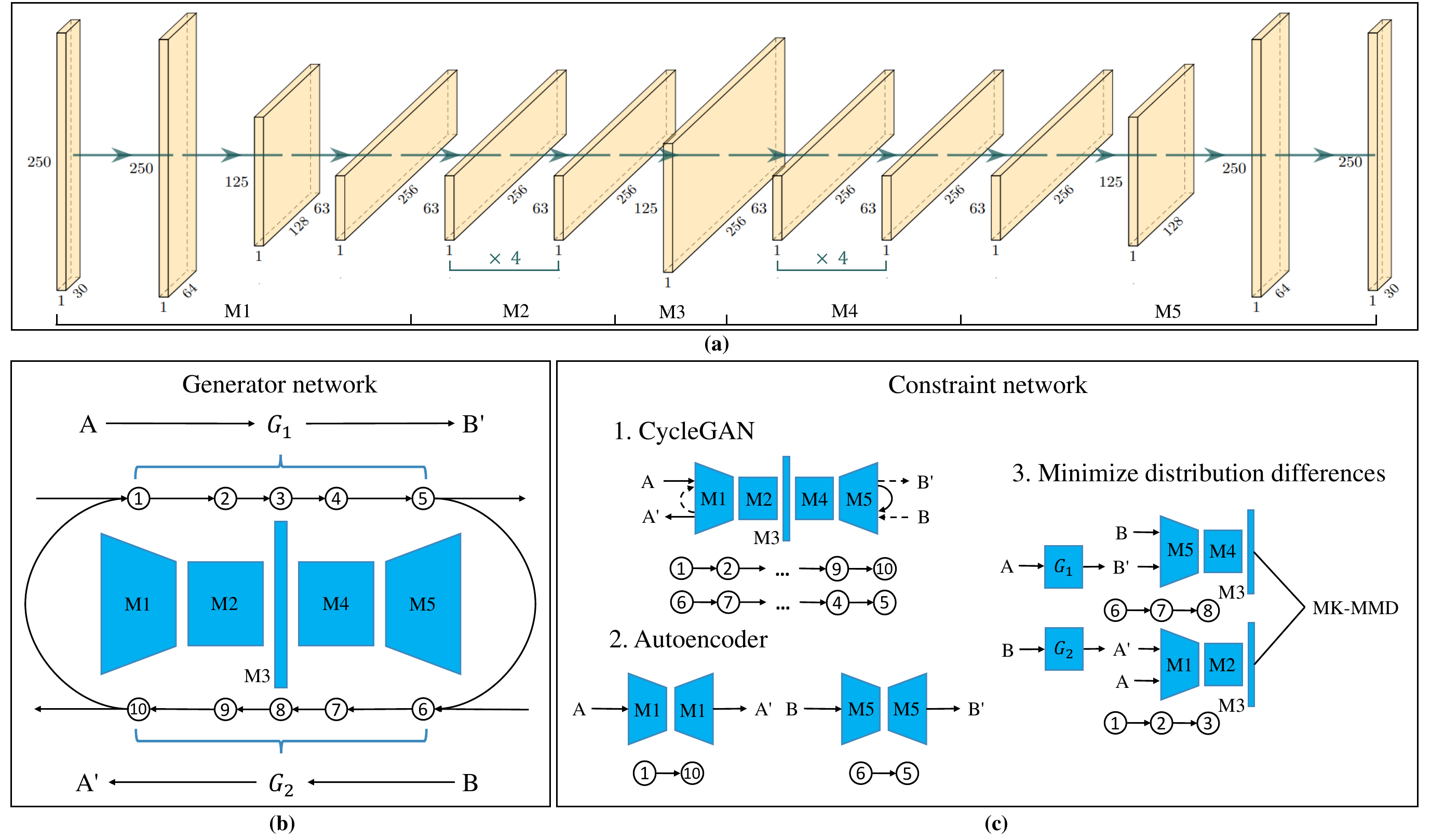}
\caption{Model architecture and modules. (a) The generator model framework in BCGGAN and the corresponding modules of each unit. (b) Since the two data conversion network models are entirely consistent and symmetrical, one network diagram was used to provide the data flow direction representing the data transformation mode. Numbers 1 to 10 are data flow steps. For example, in process 1, data are input from the left, and after extracting features of $\boldsymbol{Module1}$ ($\boldsymbol{M1}$), it is output from the right. In process 10, data are input from the right, and after extracting features of $\boldsymbol{M1}$, it is output from the left. (c) Three sub-networks were designed for different tasks. $\boldsymbol{A}$ denotes clean EEG data, $\boldsymbol{B}$ denotes BCG-corrupted EEG data, $\boldsymbol{A'}$ denotes BCGGAN-cleaned EEG data, and $\boldsymbol{B'}$ denotes BCGGAN-corrupted EEG data.}
\label{FIG:2}
\end{figure*}

This paper proposes a GAN-based method (BCGGAN) to remove the BCG artifact of simultaneous EEG-fMRI. For the GAN framework (Fig.~\ref{FIG:1}b) of this task, it is easy to build powerful discriminators but challenging to obtain reliable generators. To solve this problem, a general modular training strategy is proposed: decompose the network into multiple modules, and use multiple constraints to train the modules (Fig.~\ref{FIG:2}). In this manner, we hope to improve the local representation ability of the network model, thereby improving its overall performance and obtaining a reliable generator.  More details are described as follows.

\begin{figure}[t]
\includegraphics[width=\linewidth]{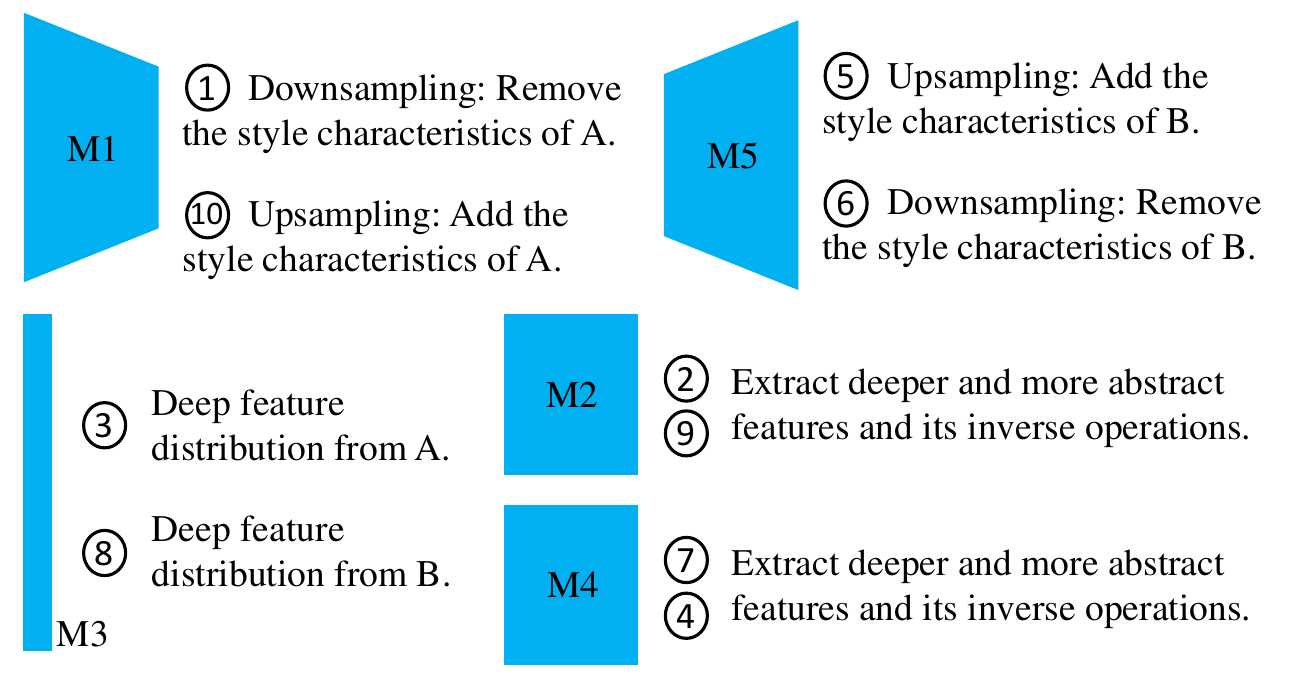}
\caption{The functions expected to be successfully achieved by the modules after training.}
\label{FIG:3}
\end{figure}

In the paper, CycleGAN (\cite{zhu2017unpaired}) is used as the basic framework (section~\ref{CycleGAN}), which contains two GANs. Each generator network is decomposed into five independent functional modules, and two constraint networks with specific constraints (section~\ref{Autoencoder},~\ref{Minimize distribution differences}) composed of different functional modules are designed (Fig.~\ref{FIG:2}). The same modules that form different networks share parameters. To improve the performance of the entire generator network model, better independent module parameters can be obtained through the synchronous training of the generator network and the constraint networks. According to the structure of CycleGAN, in the generator network ($\boldsymbol{G_1}$) of clean EEG data ($\boldsymbol{A}$) to BCGGAN-corrupted EEG data ($\boldsymbol{B'}$), the network is decomposed into five functional modules, which are the downsampling module ($\boldsymbol{M1}$), the middle feature module ($\boldsymbol{M3}$), the upsampling module ($\boldsymbol{M5}$), and two feature transform modules ($\boldsymbol{M2}$, $\boldsymbol{M4}$). The generator network ($\boldsymbol{G}_{\boldsymbol{2}}$) of BCG-corrupted EEG data ($\boldsymbol{B}$) to BCGGAN-cleaned EEG data ($\boldsymbol{A'}$) is also composed of five functional modules, which are the downsampling module ($\boldsymbol{M5}$), the middle feature module ($\boldsymbol{M3}$), the upsampling module ($\boldsymbol{M1}$), and two feature transform modules ($\boldsymbol{M4}$, $\boldsymbol{M2}$). $\boldsymbol{Mi}$ in $\boldsymbol{G}_{\boldsymbol{1}}$ has the same architecture as in $\boldsymbol{G}_{\boldsymbol{2}}$, but different parameter weights after training. Finally, in joint training, each decomposed module is expected to gradually feature the specific functions mentioned in Fig.~\ref{FIG:3}.

In the complex task of removing the BCG artifact, it is a considerable challenge to keep training in line with expectations. The several constraint networks are trained to assist in training the generator network, adding further constraints to correct the training direction. The modular training strategy can be highly convenient to guide the training of the modules by combining the excellent viewpoints to improve the overall performance of the network model.

\subsubsection{CycleGAN}
\label{CycleGAN}
The GAN framework consists of two competing networks. The first network is the discriminator, which tries to distinguish the generated data from the real data. The second network is the generator, which is trained to generate as realistic data as possible to deceive the discriminator. After a minimax game, the discriminator gradually improves the discrimination effect, while the generator can also generate progressively high-quality data. In the end, the high-performance discriminator cannot correctly classify the generated data, which shows that the generated data is almost identical to the real data. In this way, BCG-corrupted EEG data is input to generate clean EEG data.

The CycleGAN framework consists of two GANs with the same structure and used for data conversion from $\boldsymbol{A}$ to $\boldsymbol{B'}$ and $\boldsymbol{B}$ to $\boldsymbol{A'}$. Compared with a single GAN, the CycleGAN adds a cycle consistency constraint($\boldsymbol{A=A'}$, $\boldsymbol{B=B'}$). Under the constraint of $\boldsymbol{\mathcal{L}_{cycle}}$, it is ensured that the information will not be excessively lost. Therefore, CycleGAN is used as the first constraint network, called $\boldsymbol{ConstraintNet_1}$ ($\boldsymbol{CN1}$), to improve the model performance. According to the task and the guidance in WGAN (\cite{arjovsky2017wasserstein}), the loss functions are fine-tuned in CycleGAN as follows:
\begin{equation}
\begin{aligned}
\mathcal{L}_{cycle}\left(G_{1}, G_{2}\right) &= \mathbb{E}_{a \sim A}\left[{\left\|G_{2}\left(G_{1}(a)\right)-a\right\|_{1}}^{2}\right] \\
&+ \mathbb{E}_{b \sim B}\left[{\left\|G_{1}\left(G_{2}(b)\right)-b\right\|_{1}}^{2}\right]
\end{aligned}
\end{equation}
\begin{equation}
\begin{aligned}
\mathcal{L}_{GAN}\left(G_{1}, D_{B}, B, A\right) &=\mathbb{E}_{b \sim B}\left[D_{B}(b)\right] \\
&+ \mathbb{E}_{a \sim A}\left[1-D_{B}\left(G_{1}(a)\right)\right]
\end{aligned}
\end{equation}

Each signal $\boldsymbol{a}$ from $\boldsymbol{A}$ has a cycle consistency of $\boldsymbol{a}$ $\rightarrow$ $\boldsymbol{G_1}$($\boldsymbol{a}$) $\rightarrow$ $\boldsymbol{G_2}$ ($\boldsymbol{G_1}$($\boldsymbol{a}$)) $\approx$ $\boldsymbol{a}$, and so does $\boldsymbol{b}$. $\boldsymbol{\mathcal{L}_{GAN}(G_1, D_B, B, A)}$ does not use log operations, which makes the training more stable (\cite{arjovsky2017wasserstein}). $\boldsymbol{D_B}$ is the discriminator of the dataset $\boldsymbol{B}$, which is used for whether the $\boldsymbol{b}$ signal is a real signal or a synthetic signal and discriminator $\boldsymbol{D_A}$ at the same time: i.e. $\boldsymbol{\mathcal{L}_{GAN}(G_2, D_A, A, B)}$. The minimax constraints and the loss function of $\boldsymbol{CN1}$ are as follows:
\begin{equation}
\begin{aligned}
\min _{G_{1}, G_{2}} & \max _{D_{A}, D_{B}}\left(\mathcal{L}_{C N 1}\right), {where} \\
\mathcal{L}_{C N 1} &=\mathcal{L}_{\text {cycle }}\left(G_{1}, G_{2}\right) \\
&+\mathcal{L}_{G A N}\left(G_{1}, D_{B}, B, A\right) \\
&+\mathcal{L}_{G A N}\left(G_{2}, D_{A}, A, B\right)
\end{aligned}
\end{equation}

Since the SNR of the EEG signals is meager, useful physiological information is easily lost during the conversion process. If some constraints on effectively retaining information are not added, it is almost impossible to maintain the most useful physiological information. Although the loss of cycle consistency in CycleGAN can make the model more stable, the experimental results show that it has average performance. Therefore, two constraints are added based on CycleGAN, and additional training is carried out on the model to achieve improved performance.

\subsubsection{Autoencoder}
\label{Autoencoder}
$\boldsymbol{ConstraintNet_2}$ ($\boldsymbol{CN2}$) as a neural network is composed of two autoencoders (AE), which are used to improve the feature extraction ability of $\boldsymbol{M1}$ and $\boldsymbol{M5}$ modules.

An autoencoder is a special neural network structure, that adjusts the weights of network parameters through training to make the input and output values as close as possible. \cite{yang2018automatic} used the AE model to remove ocular artifacts and achieved good results. However, the BCG artifact is more difficult to remove than other types of artifact. After the removal of the BCG artifact according to the method of the AE model, the data becomes distorted. Although the AE model cannot directly solve this problem, it can indeed extract the signal's features relatively well (\cite{wen2018deep}). The $\boldsymbol{M1}$ of $\boldsymbol{G_1}$ (Data flow step 1) and the $\boldsymbol{M1}$ of $\boldsymbol{G_2}$ (Data flow step 10) can form an autoencoder ($\boldsymbol{AE_A}$) for $\boldsymbol{A}$. Similarly, the $\boldsymbol{M5}$ of $\boldsymbol{G_2}$ (Data flow step 6) and the $\boldsymbol{M5}$ of $\boldsymbol{G_1}$ (Data flow step 5) can form an autoencoder ($\boldsymbol{AE_B}$) for $\boldsymbol{B}$. The loss function of $\boldsymbol{CN2}$ is formulated as follows:
\begin{equation}
\begin{aligned}
\mathcal{L}_{CN2} &=\mathcal{L}_{AE_{A}}\left(AE_{A}, A\right)+\mathcal{L}_{AE_{B}}\left(AE_{B}, B\right) \\
&=\mathbb{E}_{a \sim A}\left[\left\|AE_{A}(a)-a\right\|_{1}^{2}\right] \\
&+\mathbb{E}_{b \sim B}\left[\left\|AE_{B}(b)-b\right\|_{1}^{2}\right]
\end{aligned}
\end{equation}

Under the synchronous training of $\boldsymbol{CN2}$, $\boldsymbol{M1}$ and $\boldsymbol{M5}$ can decode and encode signals more effectively in the up-down sampling stage.

\subsubsection{Minimize Distribution Differences}
\label{Minimize distribution differences}
$\boldsymbol{ConstraintNet_3}$ ($\boldsymbol{CN3}$) is designed to further train the $\boldsymbol{M2}$, $\boldsymbol{M3}$, $\boldsymbol{M4}$ modules.

As established in the work of \cite{yosinski2014transferable}, the abstract features learned by different layers of deep neural network will be different. As the number of network layers deepens, the features are increasingly more targeted at task-related features. Based on this idea, \cite{long2015learning} proposed the use of multi-kernel maximum mean discrepancies (MK-MMD) to measure the distribution difference of features extracted from deep layers in the neural network, and to maintain task-related features by reducing MK-MMD.

Subjects were asked to stay at rest while data were collected outside the scanner room and by the scanner. Although the noise is different for the two signals, the task-related features should be similar because the subjects are in a resting state. There are two constraints in $\boldsymbol{CN3}$: one is to maintain the similarity of task-related features of the EEG signals with resting-state collected in two different environments; the other is to perform a cyclic consistency loss on this feature to further improve stability.

For the data $\boldsymbol{A}$ and $\boldsymbol{B}$, four features are output for comparison: ${\boldsymbol{\varphi_{1}(a)}}$; $\boldsymbol{\varphi_{2}(b)}$; $\boldsymbol{\varphi_{1}({G}_{\mathbf{2}}(b))}$; $\boldsymbol{\varphi_{2}({G}_{\mathbf{1}}(a))}$. The loss function of $\boldsymbol{CN3}$ is defined as follows:
\begin{equation}
\begin{aligned}
\mathcal{L}_{MKMMD}\left(f_{1}, f_{2}\right) =\mathbb{E}\left[\left\|\mathbb{E}_{p}\left[\phi_{\kappa}\left(f_{1}\right)\right]-\mathbb{E}_{q}\left[\phi_{\kappa}\left(f_{2}\right)\right]\right\|_{\mathcal{H}_{\kappa}}^{2}\right]
\end{aligned}
\end{equation}
\begin{equation}
\begin{aligned}
\mathcal{L}_{CN3} &=\mathcal{L}_{MKMMD}\left(\varphi_{1}(a), \varphi_{2}(b)\right) \\
&+\mathcal{L}_{MKMMD}\left(\varphi_{1}(a), \varphi_{2}\left(G_{1}(a)\right)\right) \\
&+\mathcal{L}_{MKMMD}\left(\varphi_{2}(b), \varphi_{1}\left(G_{2}(b)\right)\right),
\end{aligned}
\end{equation}
where $\boldsymbol{f_1}$, $\boldsymbol{f_2}$ represent two different features, $\boldsymbol{\phi_{\kappa}(\,\cdot\,)}$ is the corresponding mapping function associated with the kernel, and ${\boldsymbol{\varphi_{1}(\,\cdot\,)}}$, ${\boldsymbol{\varphi_{2}(\,\cdot\,)}}$ are $\boldsymbol{M1(M2(M3(\,\cdot\,)))}$ and $\boldsymbol{M5(M4(M3(\,\cdot\,)))}$, respectively (corresponding to data flow step 1, 2, 3 and 6, 7, 8). The MK-MMD distance is defined as the distance between the average embeddings of two probability distributions $\boldsymbol{p}$ and $\boldsymbol{q}$ in a reproducing kernel Hilbert space (RKHS) endowed with kernel $\boldsymbol{\kappa}$.

\subsection{Training Settings}
\label{Training settings}
In BCGGAN, CycleGAN is used as $\boldsymbol{CN1}$, which is improved by combining it with some existing improvement methods (such as WGAN), and $\boldsymbol{CN2}$ and $\boldsymbol{CN3}$ are used to train the specified module further to improve the feature extraction ability of the module. The two datasets obtained in the data preprocessing stage are separately trained, and different models are created for each subject.

The two generator ($\boldsymbol{G_1, G_2}$) networks' structures are the same in BCGGAN. Therefore, the following takes $\boldsymbol{G_1}$ as an example to illustrate the model architecture. As depicted in Fig.~\ref{FIG:1}a, it contains input, downsampling ($\boldsymbol{M1}$), middle feature transform ($\boldsymbol{M2, M3, M4}$), upsampling ($\boldsymbol{M5}$), and output. The $\boldsymbol{M1}$ consists of three convolutional layers: the first convolutional layer has 64 feature maps with filter size of $7 \times 1$ and stride of 1; the next two convolutional layers respectively have 128 and 256 feature maps with filter size of $3 \times 1$ and stride of 2. Both the $\boldsymbol{M2}$ and the $\boldsymbol{M4}$ consist of eight convolutional layers: all convolutional layers have 256 feature maps with filter size of $3 \times 1$ and stride of 1. The $\boldsymbol{M3}$ consists of one deconvolutional layer, which has 256 feature maps with filter size of $3 \times 1$ and stride of 2. The $\boldsymbol{M5}$ consists of three deconvolutional layers: the first two deconvolutional layers respectively have 256 and 128 feature maps with filter size of $3 \times 1$ and stride of 2; the next deconvolutional layer has 64 feature maps with filter size of $7 \times 1$ and stride of 1. For all convolutional layers in $\boldsymbol{M1 \sim 5}$, padding and rectified linear units (ReLU) activation function are applied. The training was done using an NVIDIA TITAN Xp Founders Edition graphics card with CUDA 10.1 and cuDNN v7.6, in PyTorch v1.6.0 (\cite{paszke2019pytorch}). At an iteration of 1000 epochs, the required training time for BCGGAN model is 4.58 hours. After training, the time consumed by the model to denoise a single sample is only 11.88 milliseconds. We will open source code to make it easier for readers to understand the paper and continue research.

\subsection{Validation}
\label{Validation}
\subsubsection{Dynamic Time Warping Measures}
Dynamic Time Warping (DTW) is a dynamic programming algorithm that calculates the similarity between two time series and can find patterns in the time series (\cite{berndt1994using, muller2007dynamic}). The DTW is formulated as follows:
\begin{equation}
X=\left\{x_{1}, x_{2}, \ldots x_{m_1}\right\}, 
Y=\left\{y_{1}, y_{2}, \ldots y_{m_2}\right\}
\end{equation}
\begin{equation}
D(i, j)=d_{i j}+\min \{D(i-1, j), D(i, j-1), D(i-1, j-1)\}
\label{formula}
\end{equation}
\begin{equation}
d_{i j}=\left|x_{i}-y_{j}\right|, DTW(X, Y)=D(m_1, m_2),
\end{equation}
where $\boldsymbol{X}$, $\boldsymbol{Y}$ represent two time series with length $\boldsymbol{m_1}$ and $\boldsymbol{m_2}$ respectively, and $\boldsymbol{d_{ij}}$ represents the Euclidean distance of the two points $\boldsymbol{x_{i}}$ and $\boldsymbol{y_{j}}$. When $\boldsymbol{i=0}$ and $\boldsymbol{j=0}$, $\boldsymbol{D(i,j)}$ is equal to positive infinity, and $\boldsymbol{D(m_1,m_2)}$ is calculated by the formula~\ref{formula} as the DTW distance between $\boldsymbol{X}$ and $\boldsymbol{Y}$. In this task, the two time series are the two EEG signals before and after removing BCG artifact, so $\boldsymbol{m_1=m_2}$.

We use DTW to find patterns in EEG signals (\cite{dai2020ceneegs}) and divide the method-cleaned EEG signals into three categories: BCG artifacts can be effectively removed without introducing additional noise; BCG artifacts are not completely removed; additional noise is introduced. Finally, the DTW of the different method-cleaned EEG signals and the corresponding BCG-corrupted EEG signals are calculated to compare the performance of various methods.

\subsubsection{Peak-to-Peak Ratio Measures}
Since the BCG artifact is closely related to cardiac activity, artifact removal can be determined by observing certain signal characteristics over the period corresponding to QRS complexes. Accordingly, \cite{mantini2007complete} proposed the peak-to-peak ratio (PTPR) as a quantitative performance index to calculate the residual BCG artifact in processed EEG signals. The PTPR compares the average peaks of EEG signals before and after BCG artifact removal. The value of these peaks represents specific properties of the averages of epochs retrieved around the QRS onset. The PTPR index is described by the following formula:
\begin{equation}
PTPR=\frac{\frac{1}{n} \sum_{i=1}^{n} V_{{before}, i}(f)}{\frac{1}{n} \sum_{i=1}^{n} V_{{after}, i}(f)}=\frac{\sum_{i=1}^{n} V_{{before}, i}(f)}{\sum_{i=1}^{n} V_{{after}, i}(f)},
\label{formula:PTPR}
\end{equation}
where $\boldsymbol{n}$ represents the number of EEG channels, and $\boldsymbol{V_{{before}, i}(\,\cdot\,)}$ and $\boldsymbol{V_{{after}, i}(\,\cdot\,)}$ represent the average peak value of the channel EEG before and after the artifact.

In view of the i-channel EEG signal, we extracted epochs from the EEG from 100 ms before until 900 ms after the QRS onset. The peak-to-peak amplitudes of the averaged epochs were then computed by subtracting the minimal from the maximal value in each channel and averaged over all channels. Finally, a ratio of the resulting values before and after BCG artifact removal was calculated. When the peak-to-peak ratio is greater than 1, the BCG artifact is removed. Furthermore, a larger value of this index shows a higher ability of the algorithm to remove artifacts without distorting the data (\cite{vanderperren2010removal}).

\subsubsection{Eyes-Open/Eyes-Closed Effect}
The frequency range of BCG artifact that is difficult to remove is mainly in the theta (4–8 Hz) band (\cite{allen1998identification}), but extends to the alpha (8–13 Hz) band as well (\cite{garreffa2004simultaneous,ferdowsi2018multi}), thus overlapping with the EEG signals. If the EEG information of these frequency bands is lost during the process of BCG artifact removal, this may result in meaningless output signals. Compared to the Eyes-Open state, in the Eyes-Closed state, the increase in EEG alpha (8-13Hz) power is used as one of the result indicators of the artifact correction, which is called the Eyes-Open/Eyes-Closed (EO/EC) effect (\cite{kirschfeld2005physical}). Previous research has shown that this effect applies to most subjects. This phenomenon is evident and stable in the occipital lobe (corresponding to the position of C3, C4, and Cz electrodes). Therefore, the EO/EC effect is an excellent tool to verify whether the EEG information is lost. In the work of \cite{van2016carbon}, the meaningful physiological information was determined by analyzing the EO/EC effect that is robust and insensitive to experimental conditions. Simultaneously, \cite{barry2007eeg} studied the EEG difference between Eyes-Open and Eyes-Closed resting states and obtained the same conclusion. Due to dataset limitations, \cite{mcintosh2020ballistocardiogram} were unable to verify the EO/EC effect. The present study collected both Eyes-Open and Eyes-Closed EEG data, thus the EO/EC effect could be selected as another important indicator.
\section{Results}
The visualization results, quantification indexes, and EO/EC effect were used to compare the performance of AAS-OBS, ICA, Analyzer-2, and the deep learning methods on BCG artifact removal. These evaluation indexes are analyzed in time domain and frequency domain to verify the performance of the model (\cite{dai2020ceneegs,mantini2007complete,kirschfeld2005physical}). The results showed that BCGGAN can remove the BCG artifact more effectively while retaining useful physiological information.
Since the amplitude of the BCG artifact is larger than that of the EEG signal, the BCG artifact can be clearly seen from the time domain when the EEG signal collected by simultaneous EEG-fMRI is intuitively observed. Therefore, the performance of the BCG artifact removal method can be preliminarily judged by observing the amplitude variation after removing the BCG artifact.

Fig.~\ref{FIG:4} shows an example EEG trace of one subject was visually compared in the time domain. The blue curve indicates BCG-corrupted EEG (BCE) data, and the orange curve is the signal after artifact removal by the corresponding method. The artifact removal effects of the eight methods were compared, and the global and local performance of the generated signal curves were shown. In each method, three EEG signals were displayed. Specifically, the first EEG signal corresponded to the signal from 0 to 35000 frames. The next two EEG signals were intercepted from 8000 to 8500 frames and 18500 to 19000 frames respectively. For the last EEG signal, the gray areas marked a section without obvious BCG artifacts and two sections with obvious BCG artifacts.

Fig.~\ref{FIG:5} shows the process of calculating $\boldsymbol{DTW_i}$ ($\boldsymbol{i=1,2,3}$) and the quantitative analysis results of the signal curves of various methods. In the boxplot, we have calculated the DTW values corresponding to the three types of signals from a series of typical labeled signal segments, and the DTW value range is represented by black dashed lines and arrows. Among all the methods, in the range of $\boldsymbol{DTW_1}$ values, the signal segment generated by the BCGGAN algorithm has the most number; in the range of $\boldsymbol{DTW_2}$ values, the signal segment generated by the AAS-OBS algorithm has the most number; in the range of $\boldsymbol{DTW_3}$ values, the signal generated by the AE algorithm has the most number.

Fig.~\ref{FIG:6} shows the PTPR values of the signals before and after removing the BCG artifact in various methods. The boxplot shows the median and quartile of PTPR under multiple single channels ($\boldsymbol{n}$=1 in formula \ref{formula:PTPR}) in different methods. The numbers on the boxplot are calculated PTPR results for all channels.

Fig.~\ref{FIG:7} shows nine groups of EEG signals with Eyes-Open and Eyes-Closed and the average power spectral density of various signals in the occipital lobe. The range of the alpha band is marked with gray areas to facilitate comparison.

\section{Discussion}
Here, we introduced the BCGGAN model for removing BCG artifact in simultaneous EEG-fMRI and compare its performance with other methods, focusing on how to remove the BCG artifact more effectively while retaining essential EEG information.

\subsection{Visualization analysis}
\cite{yang2018automatic} used AE to deal with ocular artifact removal and achieved excellent performance. Their work is similar to the present study in that the same unpaired signal-to-signal problem is addressed, additional reference signal are not used, and the method does not rely on hardware. However, when AE is used for BCG artifact removal, the results demonstrate poor performance and usability, mainly because the BCG artifact is more unstable and complex than an ocular artifact (\cite{srivastava2005ica,mantini2007complete}).
\begin{figure*}[!b]
\includegraphics[width=\textwidth]{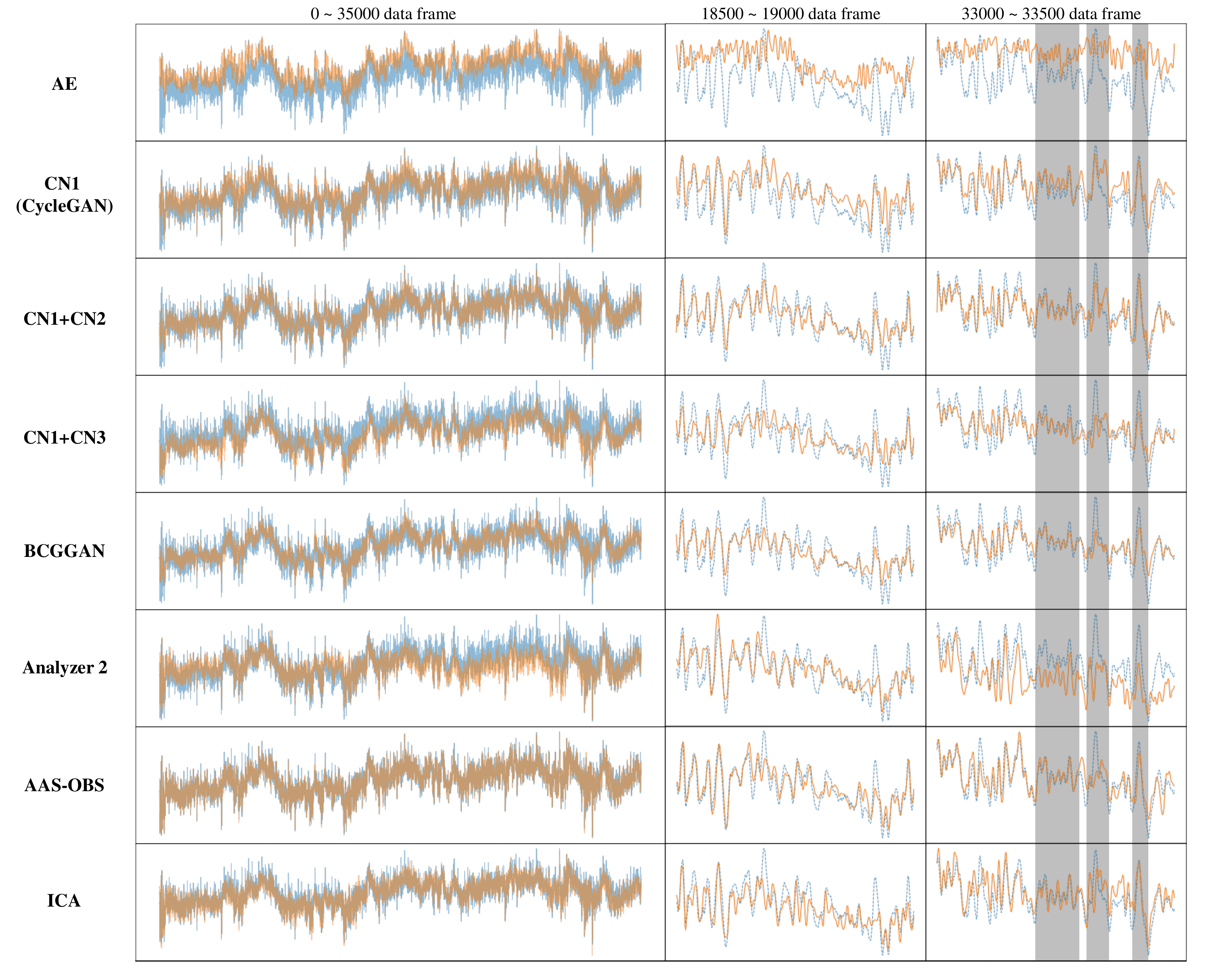}
\caption{The performance of various methods on BCG artifact removal. The blue curve indicates BCG-corrupted EEG (BCE) data, and the orange curve is the signal after artifact removal by the corresponding method. The left side consists the output result for a long time period, and the right side is the output result of two different local periods to observe the global and local denoising performance.}
\label{FIG:4}
\end{figure*}

In order to prove the effectiveness of the proposed training method, an ablation experiment is carried out to observe the changes after the deletion of $\boldsymbol{CN2}$ and $\boldsymbol{CN3}$ in BCGGAN training. From the first pair of EEG signals before (blue curve) and after (orange curve) removing the BCG artifact in Fig.~\ref{FIG:4}, the following can be found: The artifact-removed signal on the $\boldsymbol{CN1}$ model without any new constraints is more stable than the AE (\cite{yang2018automatic}). Nonetheless, it can be seen that the artifact removal effect of the $\boldsymbol{CN1}$ model is lacking. The overlap rate of the two curves is high, and some extreme peaks in the blue curve are not well-removed. After adding $\boldsymbol{CN2}$ based on $\boldsymbol{CN1}$, the artifact removal effect is slightly improved. Some extreme peaks in the blue curve have different degrees of decline in the corresponding positions in the orange curve. After adding $\boldsymbol{CN3}$ based on $\boldsymbol{CN1}$, the artifact removal effect is enhanced, but the stability of artifact-removed signal decreases. Whether the signal segment contains obvious BCG artifacts, the orange curve has a great drop compared to the blue curve. The BCGGAN model that combines $\boldsymbol{CN1}$, $\boldsymbol{CN2}$, and $\boldsymbol{CN3}$ has a better performance than the above methods, with good artifact removal effect and stability. In the signal segment with obvious BCG artifacts, the peak amplitude is reduced, while in the signal segment without obvious BCG artifacts, the amplitude of the curve remains unchanged. The gray areas in Fig.~\ref{FIG:4} show more details of the two EEG signals. The BCG-corrupted EEG signal in the first gray area did not contain obvious BCG artifacts. Except in the $\boldsymbol{CN1}$ model, where the orange curve fluctuates more than the blue curve, other methods can keep the curve unchanged. Especially in the $\boldsymbol{CN1+CN2}$ model, the artifact-removed signal is stable without generating additional noise. The BCG-corrupted EEG signals in the next two gray areas contained obvious BCG artifacts. Only $\boldsymbol{CN1+CN3}$ model and BCGGAN model remove the two BCG artifacts, but $\boldsymbol{CN1+CN3}$ model is unstable at the first artifact, resulting in signal fluctuations. The visualization results in the details also show that the BCGGAN model has a better performance.

Next, we compared BCGGAN with three other methods. Analyzer-2 (\cite{allen1998identification}) is an artifact removal technology based on the average artifact subtraction method, which uses ECG signal to detect the BCG artifact. At present, Analyzer-2 has matured as a commercial software and is widely used to remove the BCG artifact. The ICA algorithm (\cite{benar2003quality}) is used to separate the original EEG signal into different components and removes artifact components. The AAS-OBS algorithm (\cite{niazy2005removal}) uses principal components analysis to analyze signals mixed with residual artifacts on the basis of preprocessing using the AAS algorithm, and removes the residual artifact components. Based on the existing research (\cite{shams2015comparison}) and as seen in Fig.~\ref{FIG:4}, Analyzer-2 has better performance than some previous traditional technologies such as ICA, AAS-OBS, and is an essential baseline in this paper. When using Analyzer-2 to remove artifacts, the signal has a significant downward shift in the gray areas of Fig.~\ref{FIG:4}. Signal changes over a long time period are mainly caused by the state changes of the experimental subjects (\cite{le2005preictal,kar2010eeg}). Thus, the signal before and after artifact removal should not have the abovementioned signal shift situation. Meanwhile, a BCG artifact of the Analyzer-2-cleaned EEG signal at frames 8070 to 8100 has not been removed. Comparing BCGGAN and Analyzer-2, BCGGAN can more effectively remove the BCG artifact without signal shift.
\begin{figure*}[!b]
\centering
\includegraphics[width=\textwidth]{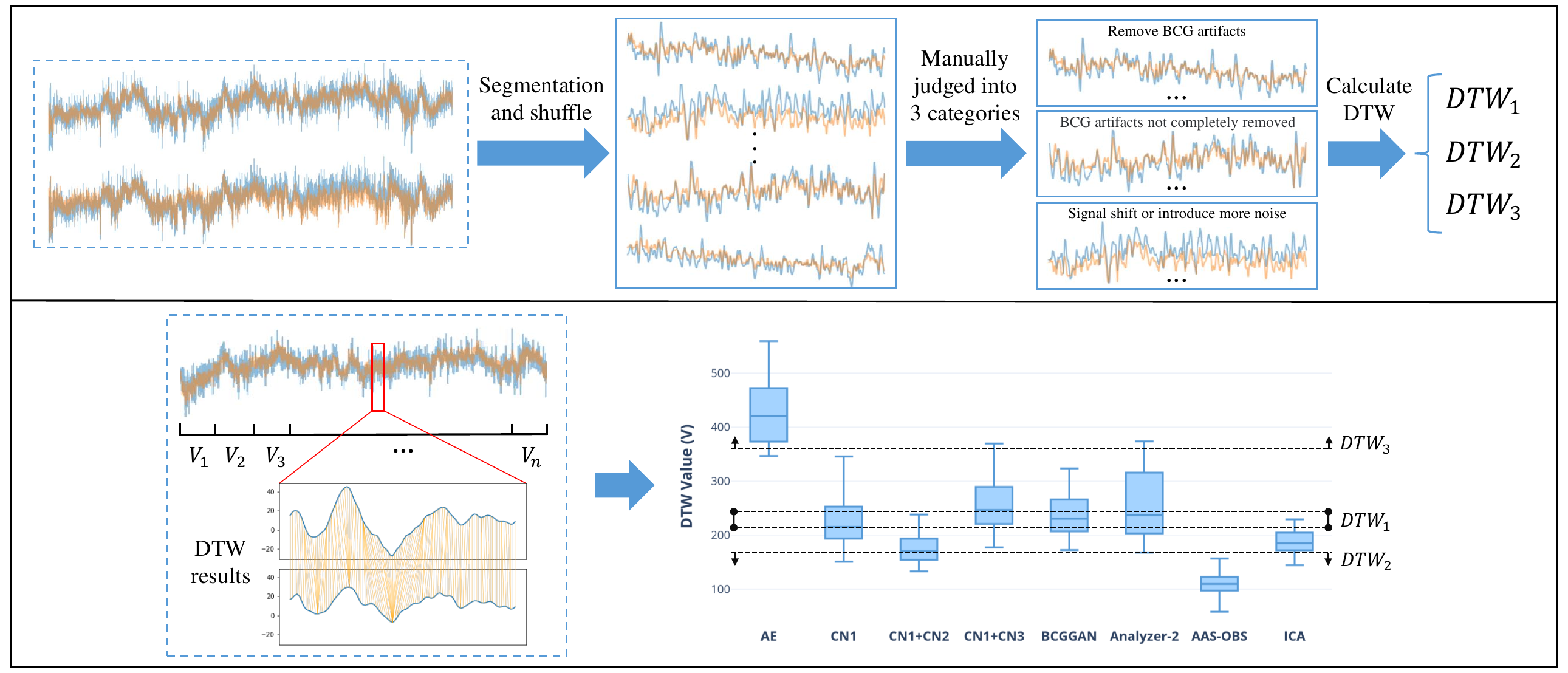}
\caption{The use of the DTW to assess the curve quality. (Top): Manually select a series of curves with obvious characteristics, which are divided into three categories: BCG artifacts can be effectively removed without introducing additional noise ($\boldsymbol{DTW_1}$ value range); BCG artifacts are not completely removed ($\boldsymbol{DTW_2}$ value range); additional noise is introduced ($\boldsymbol{DTW_3}$ value range). Calculate the range of DTW values corresponding to these three types of curves. (Bottom): After each method has removed the artifacts, calculate the DTW value ($\boldsymbol{V}$) and visualize it through the boxplot. The boxplot shows the median and quartile of $\boldsymbol{V}$ value of a signal divided into multiple data segments under different methods.}
\label{FIG:5}
\end{figure*}

After adding the $\boldsymbol{CN2}$, the model has initially acquired the ability to remove artifacts. The generated signal becomes more stable without additional fluctuations, but some artifacts are still ignored, such as the 18900 to 18950 frames in Fig.~\ref{FIG:4}. The shallow part of the network is related to some imaginal features, such as the intuitive curve shape, while the features extracted from the deeper layers are more abstract, such as some relatively abstract noises that are difficult to be described objectively (\cite{yosinski2014transferable, long2015learning}). After adding the $\boldsymbol{CN3}$, by making the deep features similar, the model can remove some different information, that is, the BCG artifact. Therefore, with the addition of these two constraints, the model can better retain the original EEG information and remove the BCG artifact. The method of removing BCG artifact based on ECG signal is difficult to deal with all the artifacts because ECG and BCG are not completely linear correlation (\cite{marino2018adaptive}), positioning deviation will lead to reduced artifact removal effect. The method of removing BCG artifact by selecting components also has a similar situation, and component selection (\cite{leclercq2009rejection,liu2012statistical}) plays a vital role in the artifact removal effect. The brain system is so complex that it is difficult to find a model to accurately describe the brain. Traditional machine learning algorithms are more interpretable and robust, but some potential information may be lost when modeling, thus affecting their performance. The BCGGAN automatically learns valuable features from data, and does not require components related to BCG artifact to be subjectively selected, which avoids potential human processing errors and improves the experimental results.

\subsection{Dynamic Time Warping Measures}
The above comparisons are all judged from a subjective human perspective, while such work requires a great amount of human resources, making it is time-consuming and inefficient. Looking at many curves for a long time to evaluate model performance is likely to cause fatigue and may lead to deviations in subsequent evaluations. This paper uses DTW to find patterns in EEG signals and evaluate method-cleaned EEG signal quality. By balancing the result accuracy and the proofreading workload, the signals are divided into data segments with a length of 1000 to calculate the corresponding DTW value. When manually selecting various types of signal segments, all signal segments must be shuffled. The continuous selection of the same type of signal tends to cause fatigue, thus blurring the selection criteria.

In the boxplot in Fig.~\ref{FIG:5}, the value ranges of different $\boldsymbol{DTW_i}$ ($\boldsymbol{i=1,2,3}$) do not overlap and have apparent intervals. Therefore, using DTW to find signal patterns is effective. In the method where $\boldsymbol{V}$ is within the range of $\boldsymbol{DTW_2}$ value, the method often has poor artifact removal effects. The signal does not change much prior to and after artifact removal, leading to a low $\boldsymbol{V}$ value. In the method where $\boldsymbol{V}$ is within the range of $\boldsymbol{DTW_3}$ value, the curve is often unstable, and the signal after artifact removal is shifted, leading to an immense $\boldsymbol{V}$ value. The proposed BCGGAN is the best method that features stable performance on different data segments, and its $\boldsymbol{V}$ is close to the value of $\boldsymbol{DTW_1}$. Although the median of the $\boldsymbol{V}$ value of Analyzer-2 is also within the range of $\boldsymbol{DTW_1}$ value, data shift occurs in some signal segments, resulting in part of the $\boldsymbol{V}$ being much larger than $\boldsymbol{DTW_1}$ value. Compared with manual evaluation, the DTW value can more intuitively display the comparison results of multiple methods in the time domain.

\subsection{Peak-to-Peak Ratio Measures}
The EEG-fMRI data obtained with open and closed eyes in the resting-state is not enough to analyze the quality of the generated curve. For example, when reviewing the DTW value, while the $\boldsymbol{CN1}$ curve has poor performance in removing artifacts, it also fluctuates in some data segments, which leads to $\boldsymbol{V}$ being close to $\boldsymbol{DTW_1}$. In this section, the peak-to-peak ratio is used to further evaluate the amount of residual BCG artifact after applying the different cleaning methods. When the PTPR is greater than 1, the BCG artifact is removed. Furthermore, a larger value of this index shows a higher ability of the algorithm to remove artifacts without distorting the data (\cite{vanderperren2010removal}).

\begin{figure}[t]
\includegraphics[width=\linewidth]{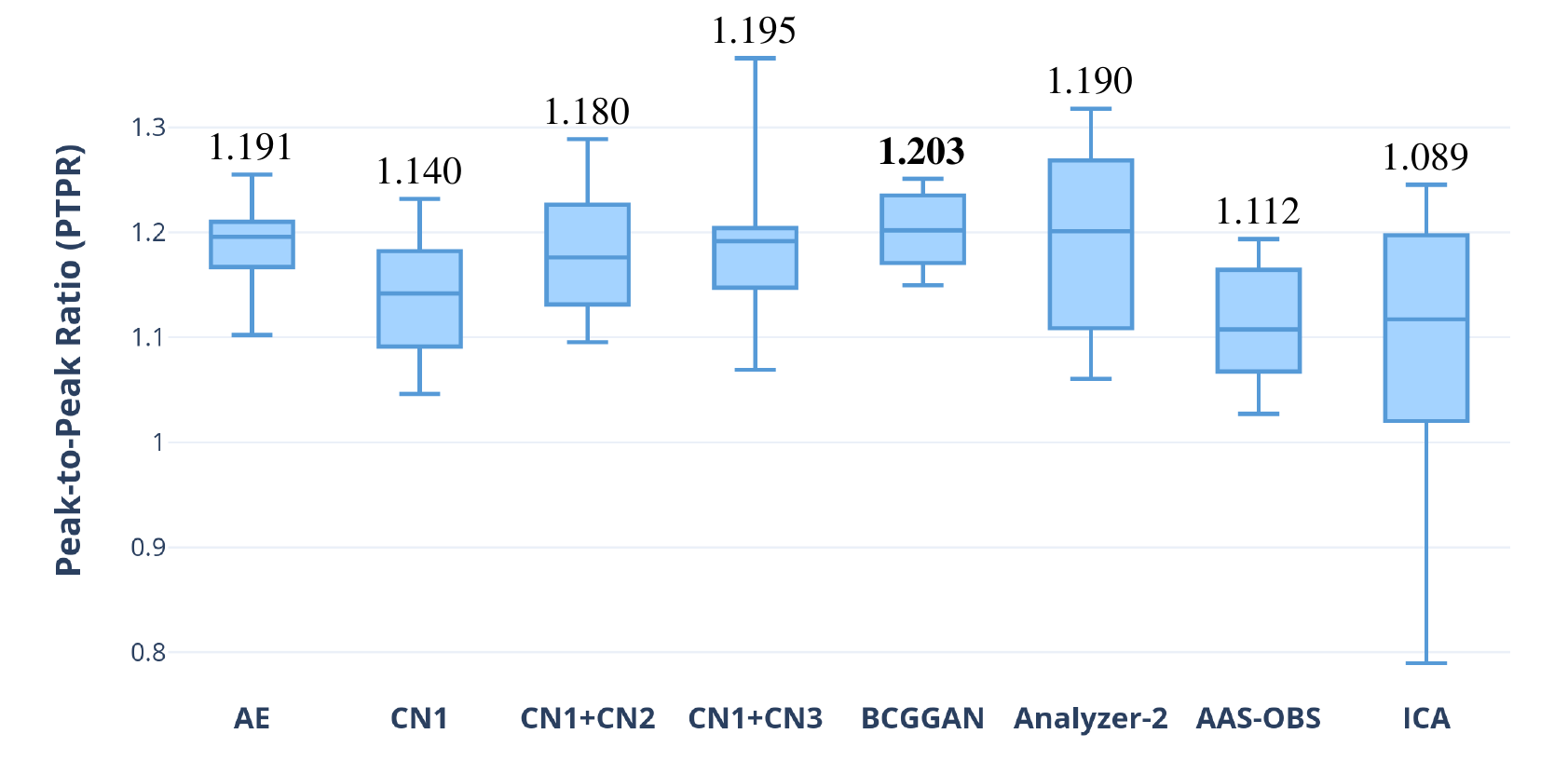}
\caption{The PTPR of various methods for BCG artifact removal.}
\label{FIG:6}
\end{figure}

As shown in Fig.~\ref{FIG:6}, BCGGAN is relatively stable under several combinations based on GAN architecture, and has satisfactory performance in removing artifacts on all channels, while its PTPR is the highest among all methods. The PTPR of $\boldsymbol{CN1}$ is the lowest in the GAN framework, and the results are consistent with the visualization results, which is reflected in the poor artifact removal performance. After adding $\boldsymbol{CN2}$ based on $\boldsymbol{CN1}$, PTPR is improved to a certain extent, but its artifact removal performance is still not good enough. After adding $\boldsymbol{CN3}$ based on $\boldsymbol{CN1}$, the value of PTPR becomes higher, while it also turns more unstable, and its performance varies significantly on different channels. Among the traditional machine learning algorithms, it is evident that Analyzer-2 has the best performance, however, there are still significant differences on various channels. The low value of PTPR of the AAS-OBS and ICA algorithms indicates that these two algorithms have poor performance in removing artifacts. The PTPR value of the ICA algorithm on some channels is less than 1, meaning that the artifacts on partial channels have not been removed.

The PTPR value of the $\boldsymbol{AE}$ and $\boldsymbol{CN1+CN2}$ methods is higher than that of Analyzer-2. However, this is not because these two methods excel at removing artifacts, but it is the result of different degrees of data distortion. It is unreliable to use a single indicator, whether the DTW or PTPR, to evaluate signals. Nevertheless, these two indicators can complement each other. It is possible to see the signal quality and whether the signal fluctuates on a single channel from the DTW, as well as the artifact removal performance of all channels from the PTPR. The experimental results show that the BCG artifact removal model based on the GAN framework is effective. Meanwhile, in BCGGAN, after $\boldsymbol{CN2}$ and $\boldsymbol{CN3}$ synchronizing training $\boldsymbol{CN1}$, the performance is improved, surpassing that of Analyzer-2.

\subsection{EO/EC effect}
Although PTPR and auxiliary variables can be used to determine the performance of the model in BCG artifact removal, it does not necessarily mean that the EEG signals after artifact removal still retain the useful physiological information. In the worst case, in the process of artifact removal, meaningful physiological information may be removed together with the BCG artifact, resulting in distortion of valuable data (\cite{van2016carbon}).
\begin{figure*}[ht]
\includegraphics[width=\textwidth]{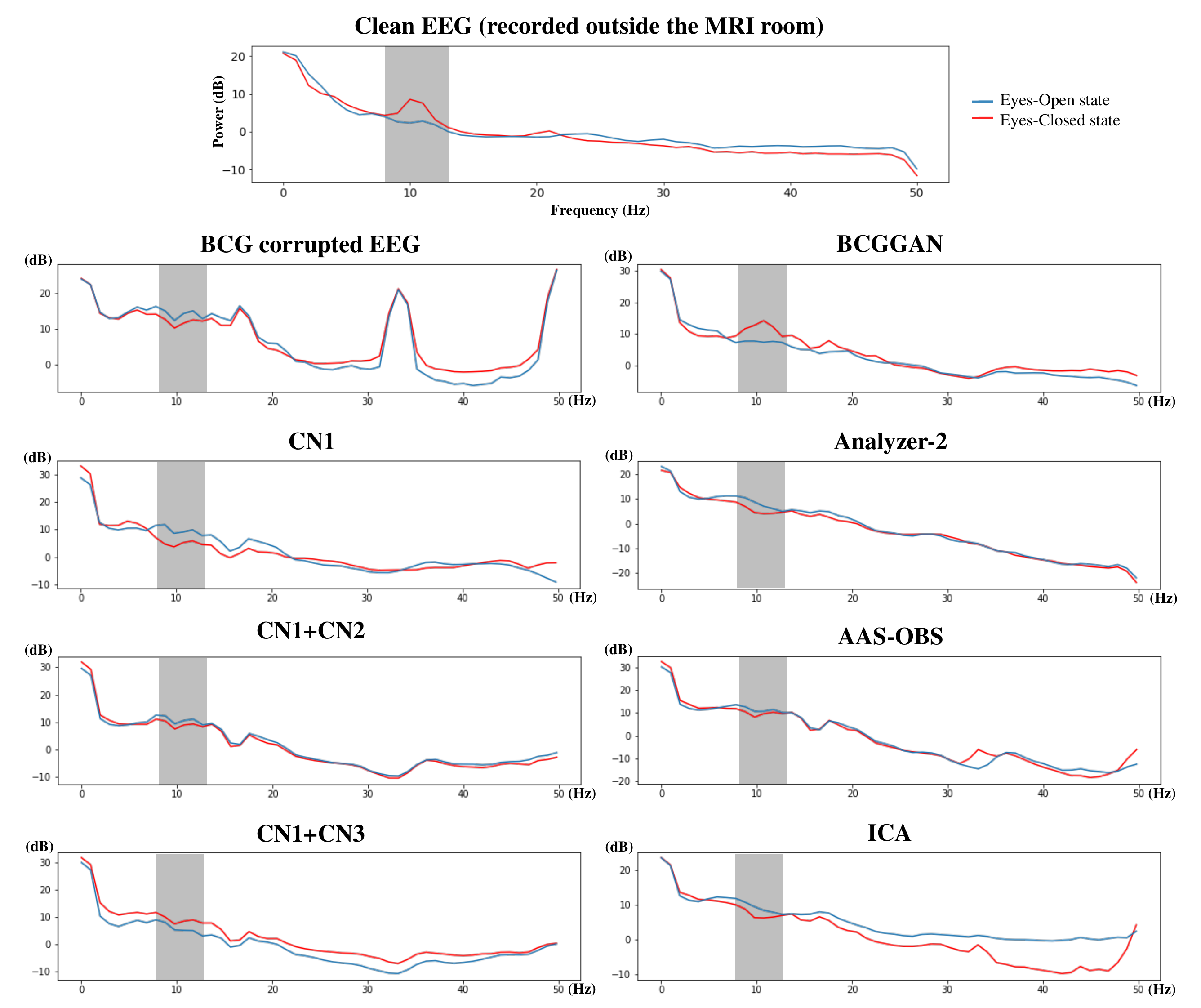}
\caption{The alpha power difference between the Eyes-Closed' (red line) and Eyes-Open (blue line) state, showing the average power spectral density of various signals in the occipital lobe.}
\label{FIG:7}
\end{figure*}

The signal difference before and after removing artifacts is clearly revealed in the EEG power spectral density (PSD). Fig.~\ref{FIG:7} shows the PSD for one subject. Seven methods that do not distort the data are further evaluated. Compared with clean EEG data (i.e., the signal data recorded outside the scanner room), the PSD of the BCG-corrupted EEG (BCE) data fluctuated wildly, which was caused by BCG artifact. After removing the artifacts from the BCE data by various methods, the PSD fluctuation of the curve in the frequency domain is smoother and relatively small. Some energy peaks caused by BCG artifact disappear in frequency bands, indicating that the artifacts have been removed. The PSD curves of BCGGAN and Analyzer-2 are the most stable ones, proving that these two methods can effectively remove the BCG artifact. As shown in each subgraph, by comparing the 'Eyes-Open state' curve and the 'Eyes-Closed state' curve, the clean EEG data shows a pronounced EO/EC effect. In the BCE data, the artifact's influence on the signal is enormous, thus it is impossible to notice the difference in alpha power between Eyes-Open and Eyes-Closed resting states. In all the methods, only the $\boldsymbol{CN1+CN3}$ model and BCGGAN model, the power of Eyes-Closed EEG data is higher than that of Eyes-Open EEG data in the alpha band, and the BCGGAN model has a more obvious EO/EC effect, the alpha power of the EEG increases in the Eyes-Closed state.

\subsection{Study limitations}
In deep learning, although the general framework of the model is determined, obtaining a usable model requires extensive testing. In training, we set several groups of the number of network layers and epochs for experiments, which have an essential impact on the research of EEG and deep learning (\cite{roy2019deep}). In the experiment, the number of $\boldsymbol{M2}$ and $\boldsymbol{M4}$ was first adjusted to 0, 2, 4, 6, 8, and 10 layers, and then gradually increased from 100 epochs to 1000 epochs at 100 intervals. This process takes a lot of time and computing resources. At the same time, due to the instability of the GAN model, sometimes the generated signal will appear chaotic. The BCGGAN model improves the stability of the GAN model as much as possible, but in a few cases, the model may crash, and the model needs to be retrained.

Another limitation of our current study is that subjects are not required to conduct specific tasks in the process of data collection. In EO/EC effect section, we use the power difference in the alpha band to prove that the signal after removing the BCG artifact still has meaningful physiological information. Indeed, experiments using EEG data with the specific tasks can further verify whether the task-related information in the EEG signals is retained. In future research, we will redesign the experimental paradigm and conduct related studies.

\section{Conclusions}
\label{Conclusions}
The removal of the BCG artifact in EEG signals recorded in the MR scanner has always been a complex task. At present, the widely used commercial software may cause data to shift in part of the data segment and cause loss of useful information. This paper designed a novel artifact removal method (BCGGAN) based on the GAN and proposed a modular training strategy to obtain a high-performance generator model. The traditional data-driven methods require an extra ECG channel to apply to BCG-corrupted EEG, however, BCG and ECG do not have a linear periodic relationship. The application of training models based on the ECG signal presents difficulties. In contrast, BCGGAN does not rely on the ECG signal, making the learning process robust and leading to promising results. On the other hand, compared to the work of \cite{van2016carbon}, our method does not require complex hardware to remove the BCG artifact. Multiple evaluations in this study proved that BCGGAN achieves better performance and is more stable than Analyzer-2. To our knowledge, our paper is the first to use the GAN model in this task while training without the use of additional hardware or reference signal, and to obtain excellent model performance at the same time.



\end{sloppypar}

\bibliographystyle{cas-model2-names}

\bibliography{cas-refs}



\end{document}